\documentclass[runningheads]{llncs}
\usepackage{graphicx}

\usepackage{tikz}
\usepackage{comment}
\usepackage{amsmath,amssymb} 
\usepackage{color}
\usepackage{orcidlink}
\usepackage{booktabs}
\usepackage{multirow}

\let\svthefootnote\thefootnote
\newcommand\freefootnote[1]{%
  \let\thefootnote\relax%
  \footnotetext{#1}%
  \let\thefootnote\svthefootnote%
}

\usepackage[accsupp]{axessibility}  

\global\long\def\modelshort{\textrm{{2G-GCN}}}
\global\long\def\datasetshort{\textrm{{MPHOI-72}}}

\begin{document}
\pagestyle{headings}
\mainmatter
\def\ECCVSubNumber{6986}  

\title{Geometric Features Informed Multi-person Human-object Interaction Recognition in Videos}

\titlerunning{Multi-person Human-object Interaction Recognition}
%
\author{Tanqiu Qiao \inst{1}\orcidlink{0000-0002-6548-0514} \and
Qianhui Men \inst{2}\orcidlink{0000-0002-0059-5484} \and
Frederick W. B. Li \inst{1}\orcidlink{0000-0002-4283-4228}\index{Li, Frederick W. B.}  \and
Yoshiki Kubotani \inst{3}\orcidlink{0000-0001-5448-7828} \and
Shigeo Morishima \inst{3}\orcidlink{0000-0001-8859-6539} \and
Hubert P. H. Shum \inst{1}$^\dag$\orcidlink{0000-0001-5651-6039}}\index{Shum, Hubert P. H.} 
\authorrunning{Qiao et al.}
%
\institute{Durham University, United Kingdom\\
\email{\{tanqiu.qiao, frederick.li, hubert.shum\}@durham.ac.uk}\\ 
\and
University of Oxford, United Kingdom 
\email{qianhui.men@eng.ox.ac.uk}\\ 
\and 
Waseda Research Institute for Science and Engineering, Japan
\email{yoshikikubotani@akane.waseda.jp, shigeo@waseda.jp}
}
\maketitle

\begin{abstract}
Human-Object Interaction (HOI) recognition in videos is important for analyzing human activity. Most existing work focusing on visual features usually suffer from occlusion in the real-world scenarios.
Such a problem will be further complicated when multiple people and objects are involved in HOIs.
Consider that geometric features such as human pose and object position provide meaningful information to understand HOIs, we argue to combine the benefits of both visual and geometric features in HOI recognition, and propose a novel Two-level Geometric feature-informed Graph Convolutional Network ($\modelshort$).
The geometric-level graph models the interdependency between geometric features of humans and objects, while the fusion-level graph further fuses them with visual features of humans and objects.
To demonstrate the novelty and effectiveness of our method in challenging scenarios, we propose a new multi-person HOI dataset ($\datasetshort$).
Extensive experiments on $\datasetshort$ (multi-person HOI), CAD-120 (single-human HOI) and Bimanual Actions (two-hand HOI) datasets demonstrate our superior performance compared to state-of-the-arts.

\end{abstract}

\section{Introduction}
\freefootnote{$\dag$ Corresponding author}
The real-world human activities are often closely associated with surrounding objects. Human-Object Interaction (HOI) recognition focuses on learning and analyzing the interaction between human and object entities for activity recognition. HOI recognition involves the segmentation and recognition of individual human sub-activities/object affordances in videos, such as drinking and placing, to gain an insight of the overall human activities \cite{morais2021learning}. Based on this, downstream applications such as security surveillance, healthcare monitoring and human-robot interactions can be developed.

Earlier work in HOI detection is limited to detecting interactions in one image \cite{gkioxari2015actions,mallya2016learning,gao2018ican}. With HOI video datasets proposed, models have been developed to learn the action representations over the spatio-temporal domain for HOI recognition \cite{qi2018learning,jain2016structural}. Notably, \cite{morais2021learning} proposes a visual feature attention model to learn asynchronous and sparse HOI in videos, achieving state-of-the-art results.

A main challenge of video-based HOI recognition is that visual features usually suffer from occlusion. This is particularly problematic in real-world scenarios when multiple people and objects are involved. Recent research has shown that extracted pose features are more robust to partial occlusions than visual features \cite{xiu2018pose,qiu2020peeking}. Bottom-up pose estimators can extract body poses as long as the local image patches of joints are not occluded \cite{cao2018openpose}. With advanced frameworks such as Graph Convolutional Networks (GCN), geometric pipelines generally perform better than visual ones on datasets with heavy occlusion \cite{das2020vpn}. Therefore, geometric features provide complementary information to visual ones \cite{bodla2021hierarchical,qiu2020peeking}.

In this paper, we propose to fuse geometric and visual features for HOI recognition in videos. Our research insight is that geometric features enrich fine-grained human-object interactions, as evidenced by previous research on image-based HOI detection \cite{zheng2020skeleton,liang2020pose}.
We present a novel Two-level Geometric feature informed Graph Convolutional Network ($\modelshort$) that extracts geometric features and fuses them with visual ones for HOI recognition in videos.
We implement the network by using the geometric-level graph to model representative geometric features among humans and objects, and fusing the visual features through the fusion-level graph.

\begin{figure}[t]
\centering
\includegraphics[keepaspectratio, scale=0.45]{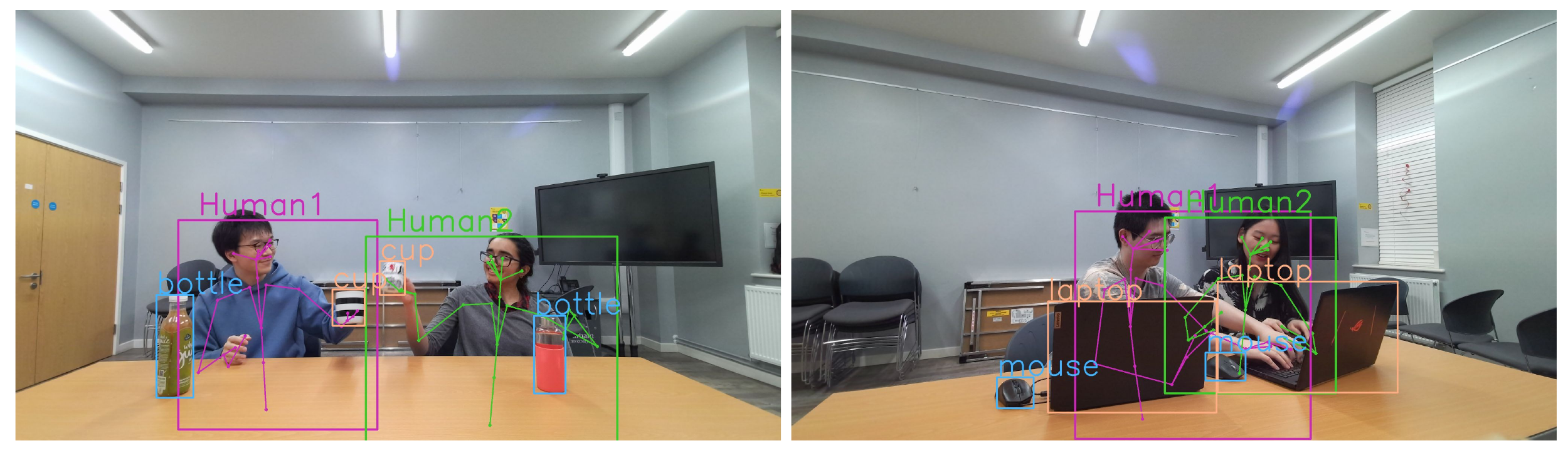}
\caption{Two examples (\textit{Cheering} and \textit{Co-working}) of our collected multi-person HOI dataset. Geometric features such as skeletons and bounding boxes are annotated.}
\label{fig:geometric_feature}
\end{figure}

To showcase the effectiveness of our model, we further propose a multi-person dataset for Human-Object Interaction (MPHOI), which closely ensembles real-world activities that contain multiple people interacting with multiple objects. 
Our dataset includes common multi-person activities and natural occlusions in daily life (Fig.~\ref{fig:geometric_feature}). It is annotated with the geometric features of human skeletal poses, human and object bound boxes, and ground-truth HOI activity labels, which can be used as a versatile benchmark for multiple tasks such as visual-based or skeleton-based human activity analysis or hybrid.

We outperform state-of-the-arts in multiple datasets, including our novel $\datasetshort$ dataset, the single-human HOI CAD-120 \cite{koppula2013learning} dataset, and the two-hand Bimanual Actions \cite{dreher2020learning} dataset. We also extensively evaluate core components of $\modelshort$ in ablation studies. Our main contributions are as follows:

\begin{itemize}
  \item We propose a novel geometry-informed $\modelshort$ network for HOI recognition in videos. The network consists of a two-level graph structure that models geometric features between human and object, together with the corresponding visual features.
  
  \item We present the novel problem of MPHOI in videos with a new $\datasetshort$ dataset, showcasing new challenges that cannot be directly resolved by existing methods. The source code and dataset are made public\footnote[1]{\url{https://github.com/tanqiu98/2G-GCN}}.
   
  \item We outperform state-of-the-art HOI recognition networks in our $\datasetshort$ dataset, the CAD-120 \cite{koppula2013learning} dataset and the Bimanual Actions \cite{dreher2020learning} dataset.
\end{itemize}

\section{Related Work}
\subsection{HOI Detection in Images} 
HOI detection aims at understanding interactions between humans and objects and identifying their interdependencies within a single image. Gupta and Malik \cite{gupta2015visual} first address the HOI detection task, which entails recognising human activities and the object instances they interact with in an image. Assigning distinct semantic responsibilities to items in a HOI process allows a detailed understanding of the present state of affairs. 
Gkioxari \textit{et al.} \cite{gkioxari2018detecting} apply an action-specific density map over target object locations depending on the appearance of an identified person to the system in \cite{gupta2015visual}. 
Multiple large-scale datasets have been presented in recent years for exploring HOI detection in images, such as V-COCO \cite{gupta2015visual}, HICO-DET \cite{chao2018learning} and HCVRD \cite{zhuang2017care}. Specifically, Mallya and Lazebnik \cite{mallya2016learning} present a simple network that fuses characteristics from a human bounding box and the entire image to detect HOIs. Gao \textit{et al.} \cite{gao2018ican} exploit an instance-centric attention module to improve the information from regions of interest and assist HOI classification. These early methods focus on visual relationships between entities in images without any potential structural relationships in HOIs.

Graph Convolutional Networks (GCN) \cite{kipf2016semi} can be used to assimilate valuable expressions of graph-structured data. Kato \textit{et al.} \cite{kato2018compositional} employ it to assemble new HOIs by using information from WordNet \cite{miller1995wordnet}. Xu \textit{et al.} \cite{xu2019learning} also exploit a GCN to model the semantic dependencies between action and object categories. Wang \textit{et al.} \cite{wang2020contextual} hypothesise that it is convenient to represent the entities as nodes and the relations as the edges connecting them in HOI. They design a contextual heterogeneous graph network to deeply explore the relations between people and objects. VSGNet \cite{ulutan2020vsgnet} refines the visual features from human-object pairs with the spatial configuration, and exploits the structural connections between pairs through graph convolution. These approaches achieve remarkable performance in image data and can provide a basis for HOI recognition in videos.

\subsection{HOI Recognition in Videos} 
HOI recognition in videos requires high-level spatial and temporal reasoning between humans and objects. Some earlier attempts apply spatio-temporal context to achieve rich-context for HOI recognition \cite{li2008key,koppula2013learning,gupta2009observing}. Recent works combine graphical models with deep neural networks (DNNs). Jain \textit{et al.} \cite{jain2016structural} propose a model for integrating the strength of spatio-temporal graphs with Recurrent Neural Networks (RNNs) in sequence learning. Qi \textit{et al.} \cite{qi2018learning} expand prior graphical models in DNNs for videos with learnable graph structures and pass messages through GPNN. Dabral \textit{et al.} \cite{dabral2021exploration} analyze the effectiveness of GCNs against Convolutional Networks and Capsule Networks for spatial relation learning. Wang \textit{et al.} \cite{wang2021spatio} propose the STIGPN exploiting the parsed graphs to learn spatio-temporal connection development and discover objects existing in a scene. Although previous methods attain impressive improvements in specific tasks, they are all based on visual features, which are unreliable in real-life HOI activities that contain occlusions between human and object entities.

\subsection{HOI Recognition Datasets} 
Multiple datasets are available to research HOI in videos for different tasks. CAD-120 \cite{koppula2013learning}, Bimanual Actions \cite{dreher2020learning}, Bimanual Manipulation \cite{KrebsMeixner2021}, \textit{etc.} are useful for single-person HOI recognition. The latter two also present bimanual HOI recognition tasks as they record human activities using both hands for object interaction. 
Something-Else \cite{materzynska2020something}, VLOG \cite{fouhey2018lifestyle}, EPIC Kitchens \cite{Damen2021RESCALING} are available for single-hand HOI recognition tasks, where the EPIC Kitchens dataset can be also used for bimanual HOI recognition since it captures both hands during cooking. 
The UCLA HHOI Dataset \cite{shu2016learning,shu2017learning} focuses on human-human-object interaction, involving at most two humans and one object. As true multi-person HOI should involve multiple humans and objects, we propose a novel MPHOI dataset that collects daily activities with multiple people interacting multiple objects. 

\subsection{Geometric Features informed HOI Analysis}
Recent research begins to employ human pose to the HOI tasks in images, which takes the advantage of capturing structured connections in human skeletons. To focus on the important aspects of interaction, Fang \textit{et al.} \cite{fang2018pairwise} suggest a pairwise body-part attention model. Based on semantic attention, Wan \textit{et al.} \cite{wan2019pose} provide a zoom-in module for extracting local characteristics of human body joints. Zheng \textit{et al.} \cite{zheng2020skeleton} introduce a skeleton-based interactive graph network (SIGN) to capture fine-grained HOI between keypoints in human skeletons and objects. 

\begin{figure}
\centering
\includegraphics[keepaspectratio, scale=0.39]{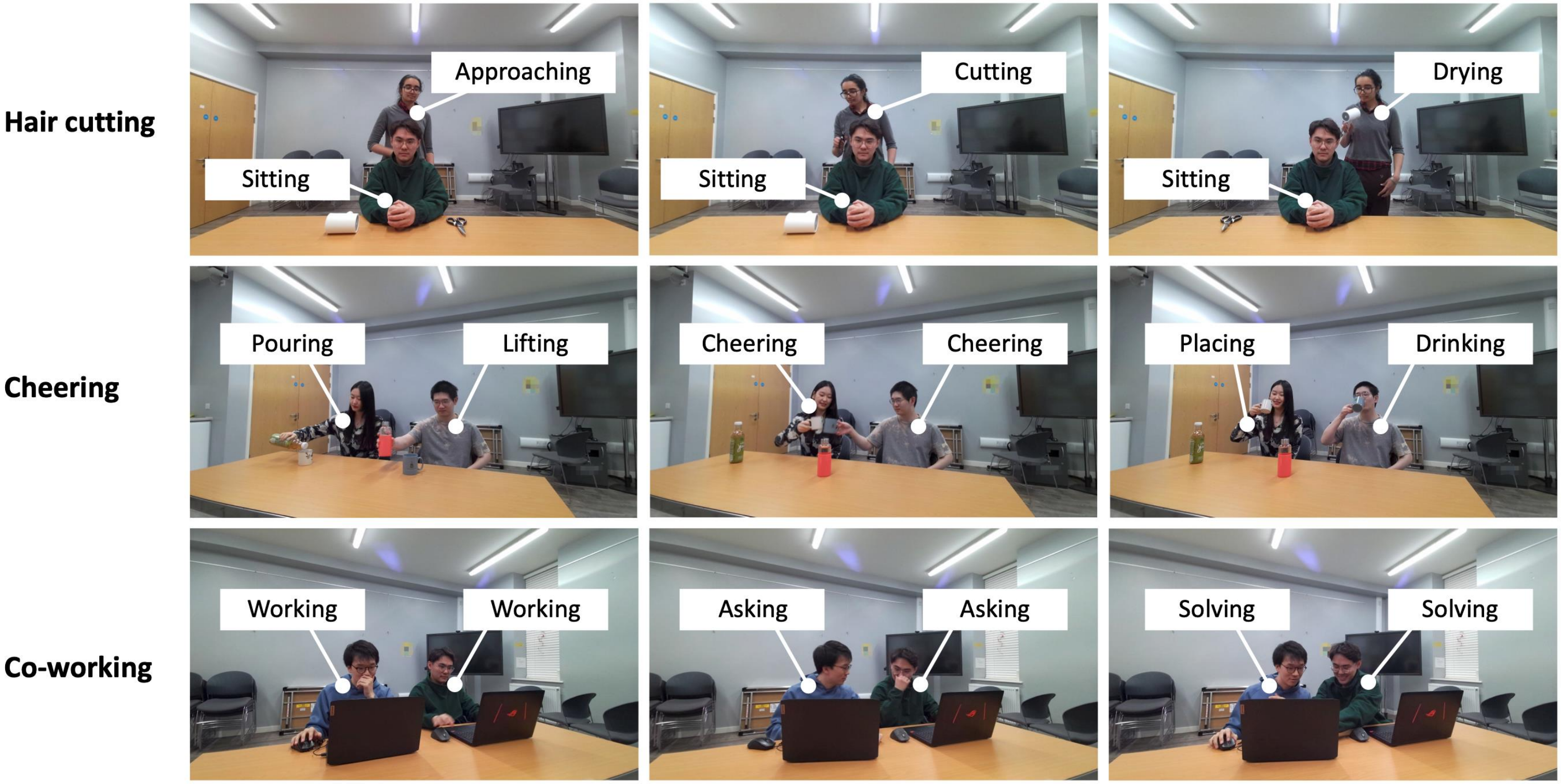}
\caption{Sample video frames of three different MPHOI activities in  $\datasetshort$.}
\label{fig:dataset}
\end{figure}

However, introducing geometric features such as the keypoints of human pose and objects to HOI learning in videos is challenging and underexplored for a few reasons. On the one hand, in a video, interaction definitions might be ambiguous, such as lift a cup vs. place a cup, approaching vs. retreating vs. reaching. These actions might be detected as the same image label due to their visual similarity. 
Videos allow the use of temporal visual cues that are not presented in images \cite{morais2021learning}. On the other hand, the model needs to consider human dynamics in the video and the shifting orientations of items in the scene in relation to humans \cite{qi2018learning}. This makes it difficult to directly extend image-based models to video that exploit the region of interest (ROI) features of human-object union \cite{dabral2021exploration}. 
We propose a novel two-level graph to refine the interactive representations; the first graph models the interdependency within the geometric key points of human and objects, and the second graph models the interdependency between the visual features and the learned geometric representations.

\section{The Multi-Person HOI Dataset ($\datasetshort$)}
We propose a HOI dataset with multi-person activities ($\datasetshort$), which is challenging due to many body occlusions among the humans and objects. We have 3 males and 2 females, aged 23-27, who are randomly combined into 8 groups with 2 people per group and perform 3 different HOI activities interacting with 2-4 objects. We also prepared 6 objects: cup, bottle, scissors, hair dryer, mouse and laptop. 3 activities = \{\textit{Cheering},  \textit{Hair cutting}, \textit{Co-working}\} and 13 sub-activities = \{\textit{Sit}, \textit{Approach}, \textit{Retreat}, \textit{Place}, \textit{Lift}, \textit{Pour}, \textit{Drink}, \textit{Cheers}, \textit{Cut}, \textit{Dry}, \textit{Work}, \textit{Ask}, \textit{Solve}\} are defined. The dataset consists of 72 videos captured from 3 different angles at 30 fps, with totally 26,383 frames and an average length of 12 seconds.

Fig.~\ref{fig:dataset} shows some sample video frames of the three activities in our $\datasetshort$ dataset, and the sub-activity label of each subject is annotated frame-wise.
The top row presents \textit{Hair cutting} from the front view, where one subject is sitting and another subject interacts with a pair of scissors and a hair dryer. Most part of the body of the subject standing at the back is invisible.
The second row presents a popular human activity, \textit{Cheering}, in which two subjects pour water from their own bottles, lift cups to cheer, and drink. The high-level occlusion exists between humans, cups and bottles during the entire activity. 
The bottom row presents \textit{Co-working}, which simulates the situation of two co-workers asking and solving questions.
Besides, we also consider distinct human sizes, skin colors and a balance of gender. These samples illustrate the diversity of our dataset.

We use Azure Kinect SDK to collect RGB-D videos with $3840\times2160$ resolution, and employ their Body Tracking SDK \cite{Kinect2022} to capture the full dynamics of two subject skeletons. Object bounding boxes are manually annotated frame-wise. For each video, we provide such geometric features: 2D human skeletons and bounding boxes of the subjects and objects involved in the activity (Fig.~\ref{fig:geometric_feature}).

\section{Two-level Geometric Features Informed Graph Convolutional Network ($\modelshort$)}
\begin{figure}
\centering
\includegraphics[height=4.7cm]{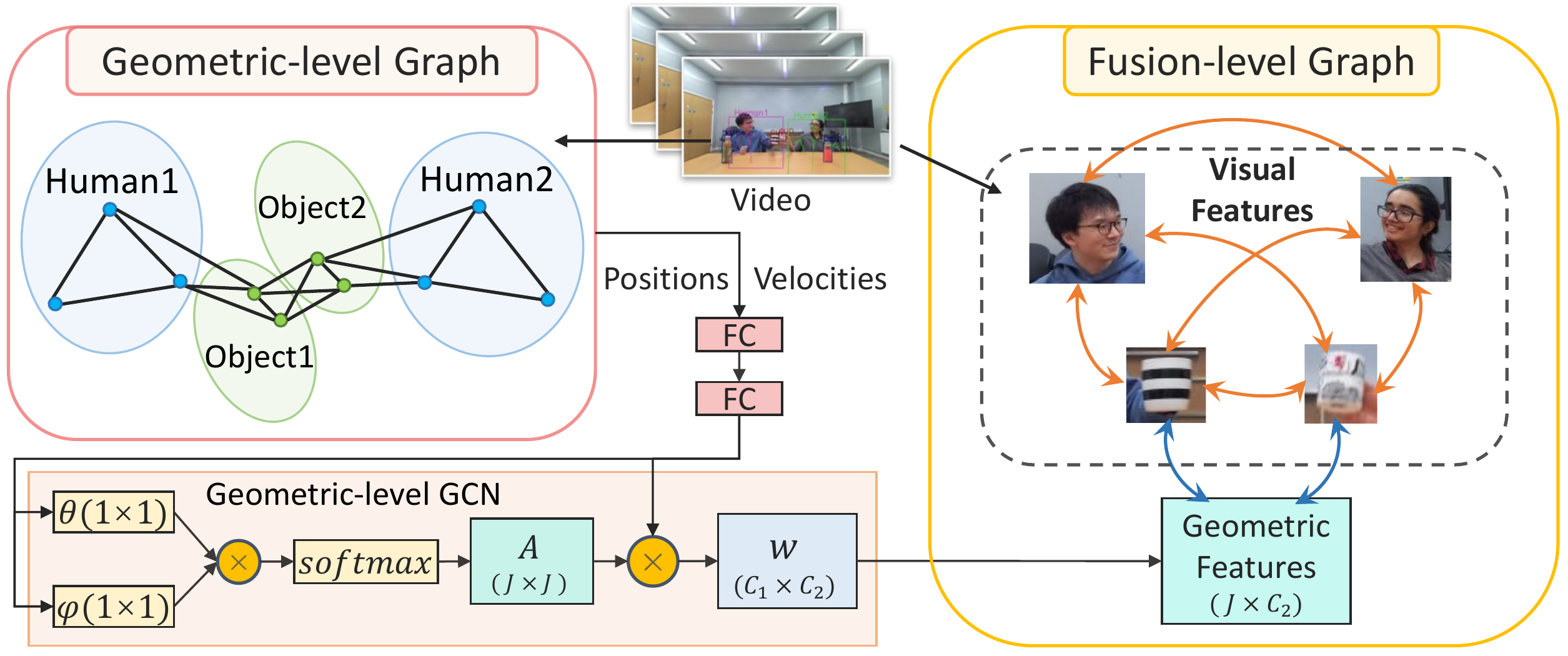}
\caption{Our $\modelshort$ framework comprises a geometric-level and a fusion-level graph.}
\label{fig:framework}
\end{figure}

To learn the correlations during human-object interaction, we propose a two-level graph structure to model the interdependency of the geometric features, known as $\modelshort$. The model consists of two key components: a geometry-level graph for modeling geometry and object features to facilitate graph convolution learning, and a fusion-level graph for fusing geometric and visual features (Fig.~\ref{fig:framework}). 

\subsection{Geometric Features}
The geometric features of humans can be represented in various ways. Human skeletons contain an explicit graph structure with joints as nodes and bones as edges. The joint position and velocity offer fine-grained dynamics in the human motion \cite{zhang2020semantics}, while the joint angle also provides spatial cues in 3D skeleton data \cite{shi2019two}. Alternatively, body shapes and how they deform during movement can be represented by surface models \cite{loper2015smpl} or implicit models \cite{Saito_2020_CVPR}. We employ human skeletons with joint position and velocity, because they are essential cues to human motion. Also, unlike body shapes, they are invariant to human appearance.

We represent human poses in an effective representation to inform HOI recognition. For human skeleton, we select specific body keypoints and denote them as a set $\mathcal{S} = \{M_t^{h,k}\}_{t=1,h=1,k=1}^{T,H,K}$, where $M_t^{h,k}$ denotes the body joint of type $k$ in human $h$ at time $t$, $T$ denotes the total number of frames in the video, $H$ and $K$ denote the total number of humans and keypoints of a human body in a frame, respectively. For a given human body keypoint $M_t^{h,k}$, we define its position as $\mathbf{p}_{t,h,k} = (x_{t,h,k}, y_{t,h,k})^T \in \mathbb{R}^2$ in 2D, and the velocity as $\mathbf{v}_{t,h,k} = \mathbf{p}_{t+1,h,k} - \mathbf{p}_{t,h,k}$, which is the forward difference of neighbour frames. In the channel of each human skeleton keypoint, we concatenate its position $\mathbf{p}_{t,h,k}$, and velocity $\mathbf{v}_{t,h,k}$ in the channel domain, forming the human geometric context $\mathbf{h}_{t,h,k}=[\mathbf{p}_{t,h,k}, \mathbf{v}_{t,h,k}] \in \mathbb{R}^4$.

As objects play a crucial role in the HOI videos, we also consider their geometric features. The two diagonal points of the object bounding box are utilised to represent the object position. We define all object keypoints as $\mathcal{O} = \{B_t^{f,u}\}_{t = 1,f=1, u=1}^{T,F,2}$, where $B_t^{f,u}$ denotes the object keypoint of type $u$ in object $f$ at time $t$. $F$ denotes the maximum number of objects in a video and $u= \{1, 2\}$ is the index of the top-left and the bottom-right points of the object bounding box, respectively. The object geometric context $\mathbf{o}_{t,f,u}=[\mathbf{p}_{t,f,u}, \mathbf{v}_{t,f,u}] \in \mathbb{R}^4$ can be obtained by the same process as the human skeleton.

\subsection{The Geometric-level Graph}
\label{sec:GG}
We design a novel geometric-level graph that involves both human skeleton and object keypoints to explore their correlations in an activity (Fig. \ref{fig:framework} left).
We use $\textbf{g}_t$ to denote a graph node with geometric features of a keypoint from either a human $h_{t,h,k}$ or a object $o_{t,f,u}$ at frame $t$.
Therefore, all keypoints of the frame $t$ are denoted by $G_{t} = (\textbf{g}_{t,1}; \cdots; \textbf{g}_{t,J})$, where $J = H\times K + F \times 2$ joints, each with 4 channel dimensions including its 2D position and velocity. This enables us to enhance the ability of GCN to capture correlations between human and object keypoints in HOI activities by learning their dynamic spatial cues. We embed $\textbf{g}_{t}$ using two fully connected (FC) layers following \cite{zhang2020semantics} as: 
\begin{equation}
    \widetilde{\textbf{g}_{t}} = \sigma(W_2(\sigma(W_1\textbf{g}_{t} + \mathbf{b}_1))+ \mathbf{b}_2) \in \mathbb{R}^{C_{1}},
    \label{equ:embedding}
\end{equation}
where $C_{1}$ is the dimension of the joint representation, $W_1 \in \mathbb{R}^{C_1 \times 4}$ and $W_2 \in \mathbb{R}^{C_1 \times C_1}$ are weight matrices, $\mathbf{b}_1$ and $\mathbf{b}_2$ are the bias vectors, and $\sigma$ is the ReLU activation function.

We propose an adaptive adjacency matrix exploiting the similarity of the geometric features in the GCN. We employ the dot-product similarity in $\widetilde{\textbf{g}_{t}}$, as it allows us to determine if and how strong a connection exists between two keypoints in the same frame $t$ \cite{wang2018non,shi2019two,zhang2020semantics}. This is a better choice for our problem comparing to other strategies, \textit{e.g.} the traditional adjacency matrix only represents the physical structure of the human body \cite{yan2018spatial} or a fully-learned adjacency matrix without supervision of graph representations \cite{shi2019two}. We represent the adjacency matrix $A_t$ with ${j_1}^{th}$ and ${j_2}^{th}$ keypoints as:
\begin{equation}
    A_t({j_1},{j_2})={\theta}(\widetilde{\textbf{g}}_{t,{j_1}})^T {\phi}(\widetilde{\textbf{g}}_{t,{j_2}}),
    \label{equ:A}
\end{equation}
where $\theta, \phi \in \mathbb{R}^{C_2}$ denote two transformation functions, each implemented by a $1 \times 1$ convolutional layer. Then, SoftMax activation is conducted on each row of $A_t$ to ensure the integration of all edge weights of a node equal to 1. We subsequently obtain the output of the geometry-level graph from the GCN as:
\begin{equation}
    Y_t =  A_{t}\widetilde{\textbf{G}}_{t}W_g,
    \label{equ:output}
\end{equation}
where $\widetilde{\textbf{G}}_{t} = (\widetilde{\textbf{g}}_{t,1}; \cdots; \widetilde{\textbf{g}}_{t,J}) \in \mathbb{R}^{J \times {C_1}}$ and $W_g \in \mathbb{R}^{{C_1}\times {C_2}}$ is the transformation matrix. The size of output is $T \times J \times C_{2}$.

\subsection{The Fusion-Level Graph}
We propose a fusion-level graph to connect the geometric features learned from GCN with visual features. Previous works on CNN-based HOI recognition in videos overemphasise visual features and neglect geometric features of humans and objects \cite{maraghi2019zero,le2020bist}. State-of-the-arts like ASSIGN \cite{morais2021learning} also exclude geometric features. In contrast, we first extract visual features for each human or object entity by ROI pooling, and then introduce the geometric output $Y_t$ from the GCN as the auxiliary feature to complement the visual representation. The feature vectors for all entities are then embedded by a two-layer MLP with ReLU activation function to the same hidden size.

A key design of the fusion-level graph is an attention mechanism to estimate the relevance of the interacted neighbouring entity. As illustrated in the fusion-level graph of Fig.~\ref{fig:framework}, each person and object denote an entity through the time, while $Y_t$ forms an additional entity joining the graph. All connections between the visual features of all humans and objects in the video are captured, represented by orange arrows. The blue arrows denote the connection between geometric and object visual features. 
Empirically, connecting the geometry-object pairs consistently performs better than applying a fully-connected graph with geometry-human connections. 
A possible reason is that humans are generally bigger in size and therefore have a larger chance of occlusion. Correlating such relatively noisy human visual and geometry features is a harder problem than the objects' equivalent.
The fusion strategy is evaluated in the ablation studies.

The attention mechanism employed in the fusion-level graph calculates a weighted average of the contributions from neighbouring nodes, implemented by a variant of scaled dot-product attention \cite{vaswani2017attention} with identical keys and values:
\begin{equation}
    \operatorname{Att}\left(q,\left\{z_{i}\right\}_{i=1 \ldots n}\right)=\sum_{i=1}^{n} \operatorname{softmax}\left(\frac{q^{T} z_{i}}{\sqrt{d}}\right) z_{i},
    \label{equ:Att}
\end{equation}
where $q$ is a query vector, $\left\{z_{i}\right\}$ is a set of keys/values vectors of size $n$, and $d$ is the feature dimension.

Once fusion-level graph is constructed, we employ ASSIGN \cite{morais2021learning} as the backbone for HOI recognition. ASSIGN is a recurrent graph network that automatically detects the structure of HOI associated with asynchronous and sparse entities in videos. Our fusion-level graph is compatible with the HOI graph structure in ASSIGN, allowing us to employ the network to predict sub-activities for humans and object-affordances for objects depending on the dataset.

\section{Experiments}
\subsection{Datasets}
We have performed experiments on our $\datasetshort$ dataset, the CAD-120 \cite{koppula2013learning} dataset and the Bimanual Actions \cite{dreher2020learning} dataset, showcasing the superior results of $\modelshort$ on multi-person, single-human and two-hand HOI recognition.

CAD-120 is widely used for HOI recognition. It consists of 120 RGB-D videos of 10 different activities performed individually by 4 participants, with each activity replicated 3 times. A participant interacts with 1-5 objects in each video. There are 10 human sub-activities (\textit{e.g.}, \textit{eating}, \textit{drinking}), and 12 object affordances (\textit{e.g.}, \textit{stationary}, \textit{drinkable}) in total, which are annotated per frame.

Bimanual Actions is the first HOI activity dataset where subjects use two hands to interact with objects (\textit{e.g.}, the left hand holding a piece of wood, while the right hand sawing it). It contains 540 RGB-D videos of 6 subjects performing 9 different activities, with each repeated for 10 times. There are totally 14 action labels for each hand and each entity in a video is annotated frame-wise.

\subsection{Implementation Details}
\subsubsection{Network Settings}
We implement 2048-dimensional ROI pooling features extracted from the 2D bounding boxes of humans and objects in the video detected by a Faster R-CNN \cite{ren2016faster} module, which is pre-trained \cite{anderson2018bottom} on the Visual Genome dataset \cite{krishna2017visual} for entity visual features.
We set the number of neurons to 64, 128 for both FC layers for the embedding and the transformation functions of Eq. \ref{equ:A} in the geometric-level graph, respectively (\textit{i.e.}, $C_{1}=64$, $C_{2}=128$).

\subsubsection{Experimental Settings}
$\modelshort$ is evaluated on two tasks: 
1) joined segmentation, and 2) label recognition given known segmentation. The first task needs the model to segment and identify the timeline for each entity in a video. The second task is a variant of the previous one, in which the ground-truth segmentation is known and the model requires to name the existing segments.
For the Bimanual Actions and CAD-120 datasets, we use leave-one-subject cross-validation to evaluate the generalization effort of $\modelshort$ in unknown subjects. On $\datasetshort$, we define a cross-validation scheme that chooses two subjects not present in the training set as the test set.

For evaluation, we report the $\mathrm{F}_{1}@k$ metric \cite{lea2017temporal} with the commonly used thresholds $k=10\%$, $ 25\%$ and $50\%$. The $\mathrm{F}_{1}@k$ metric believes each predicted action segment is correct if its Intersection over Union (IoU) ratio with respect to the corresponding ground truth is at least $k$. Since it is more sensitive to short action classes and over-segmentation errors, $\mathrm{F}_{1}$ is more adaptable than the frame-based metrics for joined segmentation and labelling issues, and was frequently adopted in prior segmentation researches \cite{lea2017temporal,farha2019ms,morais2021learning}.

With four Nvidia Titan RTX GPUs, training $\datasetshort$, CAD-120 and Bimanual Actions takes 2 hours, 8 hours and 5 days, respectively. Testing the whole test set takes 2 minutes, 3 minutes and 20 minutes, respectively.

\subsection{Quantitative Comparison}
\subsubsection{Multi-person HOIs}
In our challenging $\datasetshort$ dataset, $\modelshort$ beats ASSIGN \cite{morais2021learning} by a considerable gap (Table \ref{tab:MPHOI}). $\modelshort$ significantly outperforms ASSIGN and has smaller standard deviation values in every $\mathrm{F}_{1}$ configurations, reaching 68.6\% in $\mathrm{F}_{1}@10$ score, which is approximately 9.5\% higher than ASSIGN. 
The performance of visual-based methods such as ASSIGN is generally ineffective, since remarkable occlusions in MPHOI typically invalids visual features to HOI recognition task. The significant gaps between the results of $\modelshort$ and ASSIGN demonstrate that the application of geometric features and its fusion with visual features can motivate our model to learn stable and essential features even when significant occlusion appears in HOIs.

\begin{table}
\caption{Joined segmentation and label recognition on $\datasetshort$.\label{tab:MPHOI}}
\centering{}\resizebox{.53\textwidth}{!}{
\begin{tabular}{ccccc}
\toprule 
\multirow{2}{*}{Model} &  & \multicolumn{3}{c}{Sub-activity}\tabularnewline
\cmidrule{3-5} \cmidrule{4-5} \cmidrule{5-5} 
 &  & $\mathrm{F}_{1}@10$ & $\mathrm{F}_{1}@25$ & $\mathrm{F}_{1}@50$\tabularnewline
\cmidrule{1-1} \cmidrule{3-5} \cmidrule{4-5} \cmidrule{5-5} 
ASSIGN \cite{morais2021learning} &  & 59.1 $\pm$ 12.1 & 51.0 $\pm$ 16.7 & 33.2 $\pm$ 14.0\tabularnewline
\cmidrule{1-1} \cmidrule{3-5} \cmidrule{4-5} \cmidrule{5-5} 
$\modelshort$ &  & \textbf{68.6} $\pm$ \textbf{10.4} & \textbf{60.8} $\pm$ \textbf{10.3} & \textbf{45.2} $\pm$ \textbf{6.5}\tabularnewline
\bottomrule
\end{tabular}}
\end{table}

\begin{table*}
\caption{Joined segmentation and label recognition on CAD-120.\label{tab:cad120}}
\centering{}\resizebox{.95\textwidth}{!}{
\begin{tabular}{ccccccccc}
\toprule 
\multirow{2}{*}{Model} &  & \multicolumn{3}{c}{Sub-activity} &  & \multicolumn{3}{c}{Object Affordance}\tabularnewline
 &  & $\mathrm{F}_{1}@10$ & $\mathrm{F}_{1}@25$ & $\mathrm{F}_{1}@50$ &  & $\mathrm{F}_{1}@10$ & $\mathrm{F}_{1}@25$ & $\mathrm{F}_{1}@50$\tabularnewline
\cmidrule{1-1} \cmidrule{3-5} \cmidrule{4-5} \cmidrule{5-5} \cmidrule{7-9} \cmidrule{8-9} \cmidrule{9-9} 
rCRF \cite{sener2015rcrf} &  & 65.6 $\pm$ 3.2 & 61.5 $\pm$ 4.1 & 47.1 $\pm$ 4.3 &  & 72.1 $\pm$ 2.5 & 69.1 $\pm$ 3.3 & 57.0 $\pm$ 3.5\tabularnewline
Independent BiRNN &  & 70.2 $\pm$ 5.5 & 64.1 $\pm$ 5.3 & 48.9 $\pm$ 6.8 &  & 84.6 $\pm$ 2.1 & 81.5 $\pm$ 2.7 & 71.4 $\pm$ 4.9\tabularnewline
ATCRF \cite{koppula2016anticipating} &  & 72.0 $\pm$ 2.8 & 68.9 $\pm$ 3.6 & 53.5 $\pm$ 4.3 &  & 79.9 $\pm$ 3.1 & 77.0 $\pm$ 4.1 & 63.3 $\pm$ 4.9\tabularnewline
Relational BiRNN &  & 79.2 $\pm$ 2.5 & 75.2 $\pm$ 3.5 & 62.5 $\pm$ 5.5 &  & 82.3 $\pm$ 2.3 & 78.5 $\pm$ 2.7 & 68.9 $\pm$ 4.9\tabularnewline
ASSIGN \cite{morais2021learning} &  & 88.0 $\pm$ 1.8 & 84.8 $\pm$ 3.0 & 73.8 $\pm$ 5.8 &  & 92.0 $\pm$ 1.1 & 90.2 $\pm$ 1.8 & 82.4 $\pm$ 3.5\tabularnewline
\cmidrule{1-1} \cmidrule{3-5} \cmidrule{4-5} \cmidrule{5-5} \cmidrule{7-9} \cmidrule{8-9} \cmidrule{9-9} 
$\modelshort$ &  & \textbf{89.5} $\pm$ \textbf{1.6}  & \textbf{87.1} $\pm$ \textbf{1.8} & \textbf{76.2} $\pm$ \textbf{2.8} &  & \textbf{92.4} $\pm$ 1.7 & \textbf{
90.4} $\pm$ 2.3 & \textbf{82.7} $\pm$ \textbf{2.9}\tabularnewline
\bottomrule
\end{tabular}}
\end{table*}

\begin{table}
\caption{Joined segmentation and label recognition on Bimanual Actions.\label{tab:bimacs}}
\centering{}\resizebox{.60\textwidth}{!}{
\begin{tabular}{ccccc}
\toprule 
\multirow{2}{*}{Model} &  & \multicolumn{3}{c}{Sub-activity}\tabularnewline
\cmidrule{3-5} \cmidrule{4-5} \cmidrule{5-5} 
 &  & $\mathrm{F}_{1}@10$ & $\mathrm{F}_{1}@25$ & $\mathrm{F}_{1}@50$\tabularnewline
\cmidrule{1-1} \cmidrule{3-5} \cmidrule{4-5} \cmidrule{5-5} 
Dreher \textit{et al.} \cite{dreher2020learning} &  & 40.6 $\pm$ 7.2 & 34.8 $\pm$ 7.1 & 22.2 $\pm$ 5.7\tabularnewline
Independent BiRNN &  & 74.8 $\pm$ 7.0 & 72.0 $\pm$ 7.0 & 61.8 $\pm$ 7.3\tabularnewline
Relational BiRNN &  & 77.7 $\pm$ 3.9 & 75.0 $\pm$ 4.2 & 64.8 $\pm$ 5.3\tabularnewline
ASSIGN \cite{morais2021learning} &  & 84.0 $\pm$ 2.0 & 81.2 $\pm$ 2.0 & 68.5 $\pm$ 3.3\tabularnewline
\cmidrule{1-1} \cmidrule{3-5} \cmidrule{4-5} \cmidrule{5-5} 
$\modelshort$ &  & \textbf{85.0} $\pm$ 2.2 & \textbf{82.0} $\pm$ 2.6 & \textbf{69.2} $\pm$ \textbf{3.1}\tabularnewline
\bottomrule
\end{tabular}}
\end{table}

\subsubsection{Single-person HOIs}
The generic formulation of $\modelshort$ results in excellent performance in single-person HOI recognition. Table~\ref{tab:cad120} presents the results of $\modelshort$ with state-of-the-arts and two BiRNN-based baselines on CAD-120. Bidirectional GRU is used as a baseline in both cases: The Independent BiRNN models each entity individually (\textit{i.e.}, there are no spatial messages), but the Relational BiRNN incorporates extensive spatial relations between entities. Three previous works, ATCRF \cite{koppula2016anticipating}, rCRF \cite{sener2015rcrf} and ASSIGN \cite{morais2021learning}, are fully capable of performing this task, where ASSIGN is relatively new and can improve the scores to higher levels. For both human sub-activity and object affordance labelling, $\modelshort$ beats ASSIGN in every configuration of the $\mathrm{F}_{1}@k$ metric. Especially for the sub-activity labelling, $\modelshort$ improves 1.5\% over ASSIGN in $\mathrm{F}_{1}@10$, and more than 2\% in $\mathrm{F}_{1}@\{25, 50\}$ with lower standard deviation values. These findings demonstrate the benefits of using geometric features from human skeletons and object bounding boxes, rather than only using visual features like ASSIGN.

\subsubsection{Two-hand HOIs}
For two-hand HOI recognition on the Bimanual Actions dataset, $\modelshort$ outperforms ASSIGN \cite{morais2021learning} by about $1\%$. We compare the performance on the joined segmentation and labelling task with Dreher \textit{et al.} \cite{dreher2020learning}, ASSIGN \cite{morais2021learning} and BiRNN baselines (Table~\ref{tab:bimacs}). Dreher et al. \cite{dreher2020learning} have the worst results due to their fairly basic graph network, which ignores hand interactions and does not account for long-term temporal context. By taking into account a larger temporal context, the BiRNN baselines outperform Dreher et al. \cite{dreher2020learning}. Our $\modelshort$ has made a small improvement over ASSIGN \cite{morais2021learning}.
This is partly because the hand skeletons provided by the Bimanual Actions dataset are extracted by OpenPose \cite{cao2018openpose}, which is relatively weak on hand pose estimation.

\begin{figure}[hb]
\centering
\includegraphics[keepaspectratio, scale=0.40]{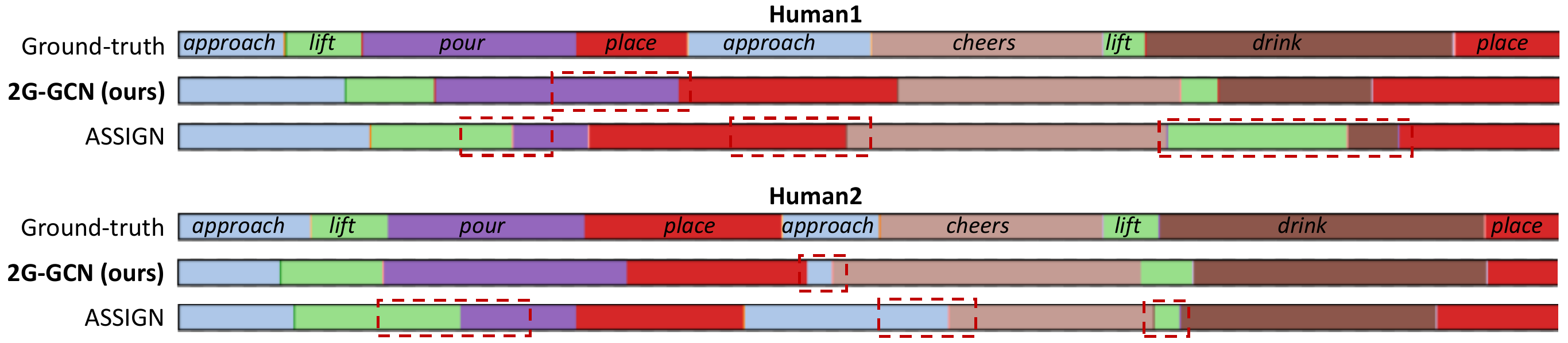}
\caption{Visualizing the segmentation and labels on $\datasetshort$ for \textit{Cheering}. Red dashed boxes highlights major segmentation errors.}
\label{fig:mphoi_vis}
\end{figure}

\begin{figure}[t]
\centering
\includegraphics[keepaspectratio, scale=0.47]{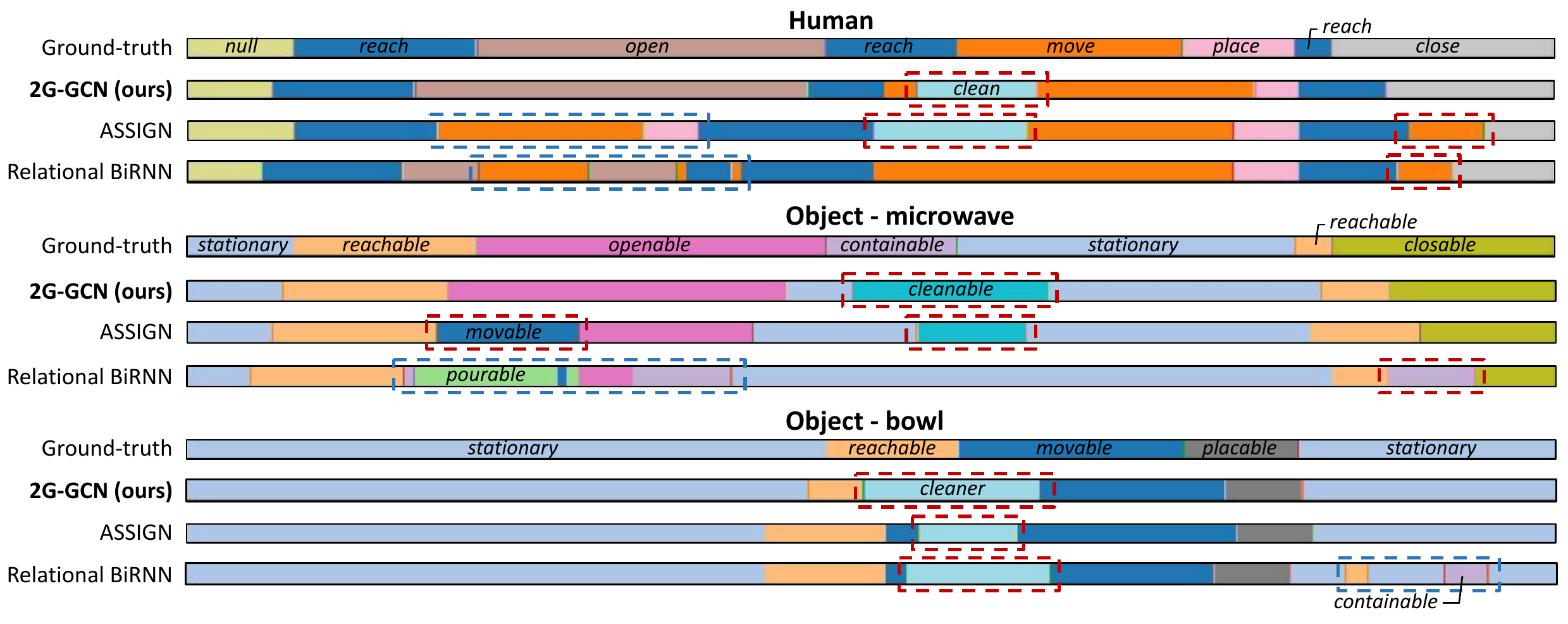}
\caption{Visualizing the segmentation and labels on CAD-120 for \textit{taking food}. Red dashed boxes highlight over-segmentation. Blue ones highlight chaotic segmentation.}
\label{fig:cad120_visualisation}
\end{figure}

\begin{figure}[t]
\centering
\includegraphics[keepaspectratio, scale=0.47]{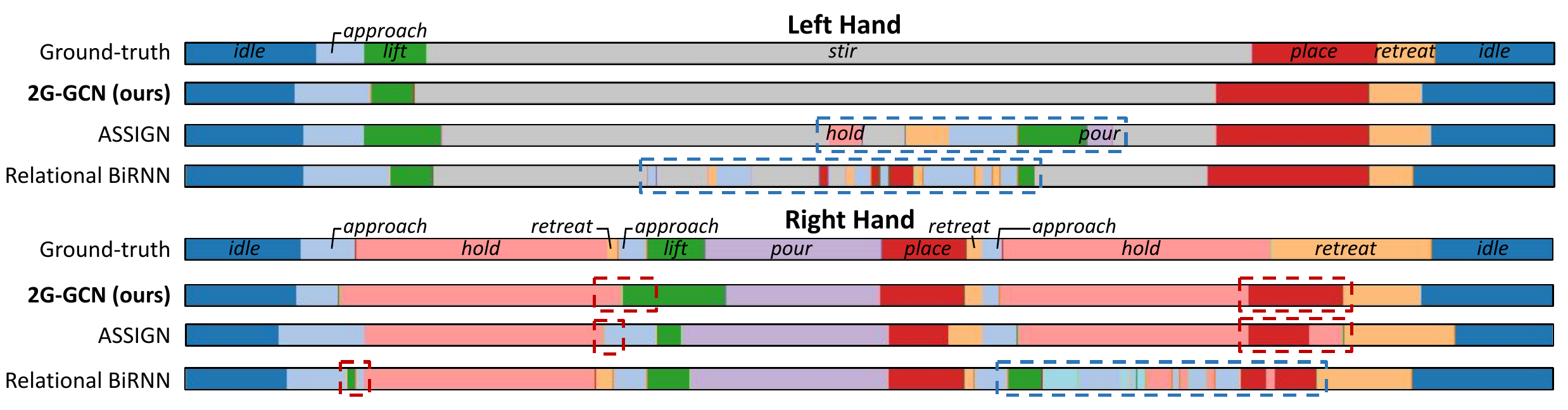}
\caption{Visualizing the segmentation and labels on Bimanual Actions for \textit{cooking}. Red dashed boxes highlight extra or missing segmentation. Blue ones highlights chaotic segmentation.}
\label{fig:bimacs_visualisation}
\end{figure}

\subsection{Qualitative Comparison}
We compare the visualization of $\modelshort$ and relevant methods on our challenging $\datasetshort$ dataset. Fig.~\ref{fig:mphoi_vis} shows an example of segmentation and labelling results with $\modelshort$ and ASSIGN \cite{morais2021learning} approaches compared with the ground-truth for a \textit{Cheering} activity. We highlight some major segmentation errors with red dashed boxes. Although both models have some errors, $\modelshort$ is generally more robust to varying segmentation period and activity progression than ASSIGN. $\modelshort$ is not particularly sensitive to the timeline of \textit{place} and \textit{approach}, while ASSIGN crashes for most of the activities.

Fig. \ref{fig:cad120_visualisation} displays an example of a \textit{taking food} activity on the CAD-120 dataset. We highlight over-segmentation with the red dashed box and chaotic segmentation with the blue dashed box. From the figure, our $\modelshort$ is able to segment and recognise both human sub-activities and object affordances more accurately than the other two models. ASSIGN \cite{morais2021learning} and Relational BiRNN fail to predict when the human opens or closes the microwave (\textit{e.g.} the \textit{open} and \textit{close} sub-activities for the human, and the \textit{openable} and \textit{closable} affordances for the microwave).

Fig. \ref{fig:bimacs_visualisation} depicts the qualitative visualization of a \textit{cooking} activity on the Bimanual Actions dataset. Here, $\modelshort$ performs outstandingly with precise segmentation and labelling results for the left hand, while ASSIGN \cite{morais2021learning} and Relational BiRNN have a chaotic performance when segmenting the long \textit{stir} action. In contrast, the right hand has more complex actions, which confuses the models a lot. $\modelshort$ generally performs better than ASSIGN, although both of them have some additional and missing segmentations. Relational BiRNN has the worst performance with chaotic segmentation errors in the \textit{hold} action.

\subsection{Ablation Studies}
The two proposed graphs in our method contain important structural information. We ablate various essential modules and evaluate them on the CAD-120 dataset to demonstrate the role of different $\modelshort$ components as shown in Table~\ref{tab:ablation}, where GG and FG denote the geometric-level graph and fusion-level graph, respectively.

\begin{table}
\caption{Ablation study on CAD-120. GG and FG denote the geometric-level graph and the fusion-level graph, respectively.  \label{tab:ablation}}
\centering{}\resizebox{0.75\columnwidth}{!}{
\begin{tabular}{clcccccc}
\toprule 
 & \multirow{2}{*}{Model} &  & \multicolumn{2}{c}{Sub-activity} &  & \multicolumn{2}{c}{Object Affordance }\tabularnewline
 &  &  & $\mathrm{F}_{1}@10$ & $\mathrm{F}_{1}@25$ &  & $\mathrm{F}_{1}@10$ & $\mathrm{F}_{1}@25$\tabularnewline
\cmidrule{1-2} \cmidrule{2-2} \cmidrule{4-5} \cmidrule{5-5} \cmidrule{7-8} \cmidrule{8-8} 
(1) & GG (w/o skeletons) \& FG &  & 87.7 & 84.9 &  & 91.0 & 88.3\tabularnewline
(2) & GG (w/o objects) \& FG &  & 88.3 & 85.6 &  & 90.4 & 88.5\tabularnewline
(3) & GG (w/o embedding) \& FG &  & 89.4 & 86.4 &  & 91.5 & 90.0\tabularnewline
(4) & GG (w/o similarity) \& FG &  & 88.7 & 85.0 &  & 90.6 & 89.0\tabularnewline
\cmidrule{1-2} \cmidrule{2-2} \cmidrule{4-5} \cmidrule{5-5} \cmidrule{7-8} \cmidrule{8-8} 
(5) & GG \& FG (w/o human-object) &  & 73.4 & 68.8 &  & 90.3 & 88.4\tabularnewline
(6) & GG \& FG (w/o object-object) &  & 88.3 & 84.5 &  & 90.9 & 88.5\tabularnewline
(7) & GG \& FG (w human-geometry) &  & 89.0 & 86.6 &  & 91.4 & 89.3\tabularnewline
\cmidrule{1-2} \cmidrule{2-2} \cmidrule{4-5} \cmidrule{5-5} \cmidrule{7-8} \cmidrule{8-8} 
(8) & $\modelshort$ &  & \textbf{89.5} & \textbf{87.1} &  & \textbf{92.4} & \textbf{90.4}\tabularnewline
\bottomrule
\end{tabular}}
\end{table}

We firstly investigate the importance for geometric features of the human and objects. The experiments in row (1) drops the human skeleton features in the geometric-level graph, while row (2) drops the object keypoint features. Row (3) explores the effect of the embedding function on geometric features. The last component we ablated is the similarity matrix used in the GCN, the result comparison between row (4) and (8) demonstrates its significance in the model.

\begin{figure}[t]
\centering
\includegraphics[height=4.0cm]{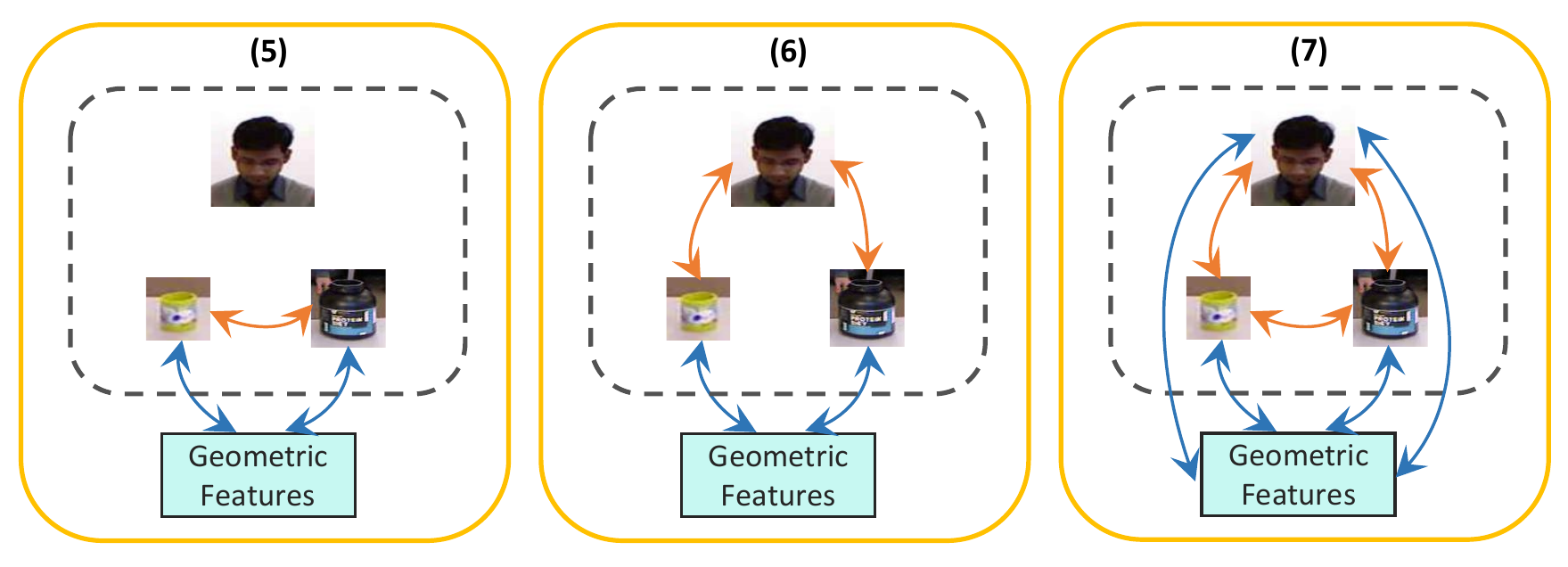}
\caption{Ablation study of the fusion-level graph. Human-object, object-object and geometry-human relations are ablated (rows (5), (6), (7) in Table~\ref{tab:ablation} respectively).}
\label{fig:ablation}
\end{figure}

We further ablate different components in the fusion-level graph as shown in Fig.~\ref{fig:ablation}. We disable the attention connection between the pair of human-object and object-object in row (5) and (6), respectively, and also supplement the human-geometry connection in row (7). The inferior results reported in row (5) and (6) verify the significance of incorporating all these pair connections in our full $\modelshort$ model.

\section{Conclusions}
We propose a two-level graph GCN for tackling HOIs in videos, which consists of a geometric-level graph using human skeletons and object bounding boxes, and a fusion-level graph fusing the geometric features with traditional visual features. We also propose a novel $\datasetshort$ dataset to enable and motivate research in multi-person HOI recognition.
Our $\modelshort$ outperforms state-of-the-art HOI recognition networks in single-person, two-hand and multi-person HOI domains.

Our method is not limited to two humans; the geometric-level graph can represent multiple humans and objects. To handle an arbitrary number of entities, a graph can be constructed by only considering the k-nearest humans and objects, allowing better generalisation \cite{mohamed2020social}.
If there are a large number of entities, to avoid handling a large fully-connected graph, we can apply an attention mechanism to learn what nodes are related \cite{shi2021sgcn}, thereby better recognising HOIs.

We found that the accuracy of skeleton joint detection can affect the quality of geometric features. In future work, we may employ some algorithms for noise handling. Skeleton reconstruction methods such as the lazy learning approach \cite{shum2013real} or motion denoising methods such as the deep learning manifold \cite{wang21spatiotemporal} would enhance the accuracy of skeleton information based on prior learning from a dataset of natural motion. One of our future directions is to employ such techniques to improve our geometric features.

Another future direction is to enrich the geometric representation of objects. While the bounding box features are powerful, it cannot represent the geometric details \cite{zhu22skeleton}. On the one hand, the rotation-equivariant detector \cite{Han_2021_CVPR} enriches the object representation with rotated bounding boxes, resulting in improved object detection performance. On the other hand, the recently proposed convex-hull features \cite{Guo_2021_CVPR} allow representing objects of irregular shapes and layouts. They could enhance our geometric feature based HOI recognition framework significantly.

\clearpage
%
%
\bibliographystyle{splncs04}
\bibliography{egbib}
\end{document}